\documentclass[10pt, conference, compsocconf]{IEEEtran}

\ifCLASSINFOpdf
\else
\fi

\usepackage{subfigure}
\usepackage{multirow}
\usepackage{booktabs}
\usepackage{graphicx}
\usepackage{textcomp,enumitem}
\usepackage{xcolor}
\usepackage{algorithm, algorithmicx, algpseudocode}

\usepackage{cite}
\usepackage{amsmath,amssymb,amsfonts}
\usepackage{lipsum}

\newcommand\blfootnote[1]{%
  \begingroup
  \renewcommand\thefootnote{}\footnote{#1}%
  \addtocounter{footnote}{-1}%
  \endgroup
}

\usepackage{url}

\begin{document}
\title{Generalizing Universal Adversarial Attacks \\Beyond Additive Perturbations}


\author{\IEEEauthorblockN{Yanghao Zhang\IEEEauthorrefmark{1},
Wenjie Ruan\IEEEauthorrefmark{1}\IEEEauthorrefmark{2}, Fu Wang\IEEEauthorrefmark{1} and
Xiaowei Huang\IEEEauthorrefmark{3}}
\IEEEauthorblockA{\IEEEauthorrefmark{1}College of Engineering, Mathematics and Physical Sciences, University of Exeter, Exeter, UK \\
\IEEEauthorrefmark{3}Department of Computer Science, University of Liverpool, Liverpool, UK\\
\IEEEauthorrefmark{1}\{yanghao.zhang,w.ruan\}@exeter.ac.uk, \IEEEauthorrefmark{1}fuu.wanng@gmail.com,
\IEEEauthorrefmark{3}xiaowei.huang@liverpool.ac.uk}
}


%


\maketitle

\begin{abstract}
The previous study has shown that universal adversarial attacks can fool deep neural networks over a large set of input images with a single human-invisible perturbation.
However, current methods for universal adversarial attacks are based on additive perturbation, which cause misclassification when the perturbation is directly added to the input images. 
In this paper, for the first time, we show that a universal adversarial attack can also be achieved via non-additive perturbation (e.g., spatial transformation). More importantly, to unify both additive and non-additive perturbations, we propose a novel unified yet flexible framework for universal adversarial attacks, called GUAP, which is able to initiate attacks by additive perturbation, non-additive perturbation, or the combination of both.
Extensive experiments are conducted on CIFAR-10 and ImageNet datasets with six deep neural network models including GoogleLeNet, VGG16/19, ResNet101/152, and DenseNet121. 
The empirical experiments demonstrate that GUAP can obtain up to 90.9\% and 99.24\% successful attack rates on CIFAR-10 and ImageNet datasets, leading to over 15\% and 19\% improvements respectively than current state-of-the-art universal adversarial attacks.  
The code for reproducing the experiments in this paper is available at \url{https://github.com/TrustAI/GUAP}.
\end{abstract}


\IEEEpeerreviewmaketitle

\blfootnote{\IEEEauthorrefmark{2} Corresponding author. \\
This work is supported by Partnership Resource Fund (PRF) on Towards the Accountable and Explainable Learning-enabled Autonomous Robotic Systems from UK EPSRC project on Offshore Robotics for Certification of Assets (ORCA) [EP/R026173/1], and the UK Dstl project on Test Coverage Metrics for Artificial Intelligence.}

\section{Introduction}

\begin{figure}[!tb]
\centering
\includegraphics[width=\columnwidth]{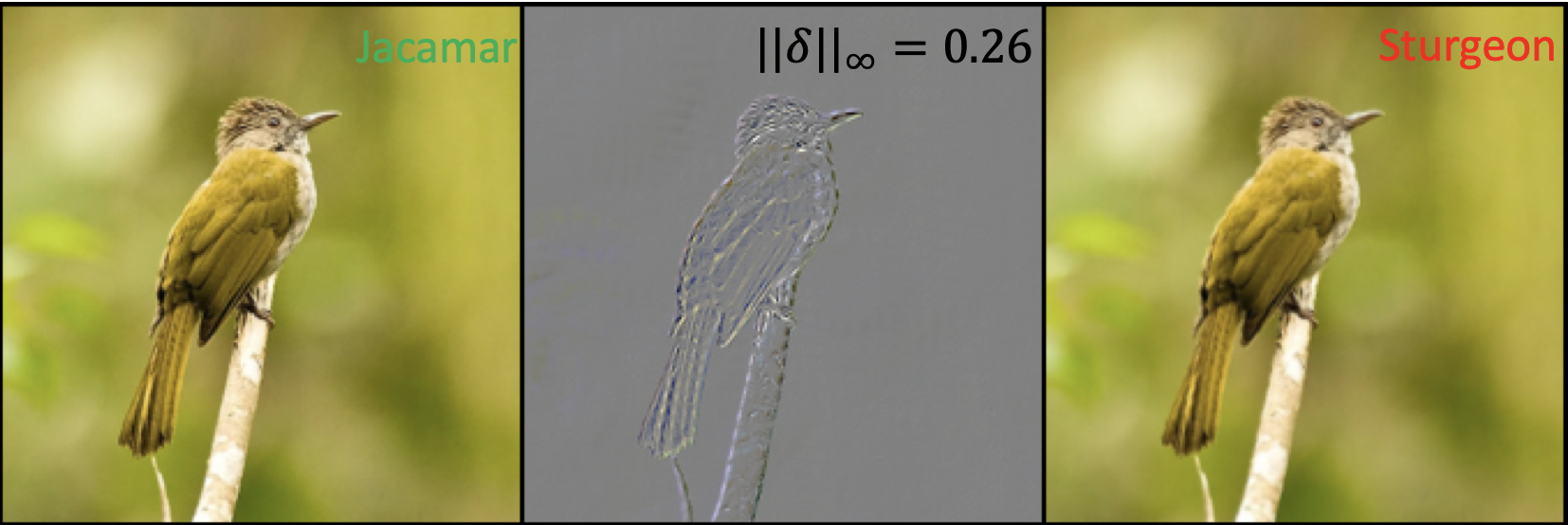}
\caption{Adversarial example generated by spatially transformation, the VGG19 neural network can be easily fooled so it incorrectly classifies the ``Jacamar'' as ``Sturgeon'', the middle column represents the intermediate perturbation scaled with the minimum of 0 and the maximum of 1.}
\label{spexample}
\end{figure}

Although deep neural networks (DNNs) have achieved huge success across a wide range of applications, yet recently some researchers demonstrated that DNNs are vulnerable to adversarial examples or attacks~\cite{SzegedyZSBEGF13,carlini2017towards,huang2019coverage,safetysurvey}. 
Adversarial examples are generated by adding small perturbations to an input, sometimes imperceptible to humans, that can enable the neural network to make an incorrect classification result~\cite{zhang2019generation,WWRHK2018,SWRHKK2018}. Taking Fig.~\ref{spexample} as an example, by adding a human-invisible perturbation, the VGG19 neural network can be easily fooled such that it incorrectly classifies the image ``Jacamar'' as ``Sturgeon''.

Thus, adversarial examples~\cite{goodfellow2014explaining,xu2020towards} have become a serious risk especially when DNNs are applied to security-critical applications such as medical record analysis~\cite{sun2018identify}, malware detection~\cite{wang2017adversary}, and autonomous vehicles~\cite{carlini2017towards}.   
Most existing adversarial attack methods focus on generating an adversarial perturbation over a specific input~\cite{carlini2017towards,zhang2019generation,goodfellow2014explaining,RHK2018}. 
These perturbations are image specific, i.e., different perturbations are generated for different inputs. 
An adversarial perturbation of this type may expose the weakness of the network within the local precinct of the original image in the input domain, but it cannot directly support the analysis of global robustness \cite{RWSHKK2019}.
In order to support this, the concept of universal adversarial perturbation is considered, which can fool a well-trained neural network on a set of, ideally all, input images. 
Moosavi-Dezfooli et al. \cite{moosavi2017universal} firstly showed the existence of the Universal Adversarial Perturbation (UAP) and presented an iterative algorithm to compute it based on a set of input images. 
Unlike UAP employing iterative method, some other works also showed that generative models can be used for crafting universal perturbation \cite{hayes2018learning,poursaeed2018generative,reddy2018nag}, with the aim to capture the distribution of adversarial perturbation and produce a higher fooling rate.

Until today, existing universal adversarial attacks are all additive (i.e., they make an image misclassified when added directly to the image) and based on $\ell_p$ norm distance to constrain the magnitude of the perturbation.
However, $\ell_p$ norm may not be a suitable measurement (for capturing the human perception) when it comes to a perturbation in terms of transformation. 
The spatially transformed adversarial example~\cite{xiao2018spatially} in Fig.~\ref{spexample} is almost “the same” to human perception but results in a large $\ell_\infty$ distance.
This observation had led to another type of adversarial perturbation, i.e. non-additive perturbation. 
Generally speaking, a non-additive perturbation can be seen as a functional transformation on the input, and hence a generalisation to the additive perturbation. 
Recently, a particular type of non-additive perturbation, i.e., spatial transformation, has increasingly attracted the attention from the community. Some researchers observed that deep neural models suffer from spatial variants of the input data while, conversely, humans are usually less sensitive for such spatial distortions~\cite{lenc2015understanding,jaderberg2015spatial}. In this regard, some pioneering works are emerged very recently, which aim to generate spatially transformed adversarial examples to fool the deep neural networks~\cite{engstrom2019exploring,xiao2018spatially}. For instance, 
Engstrom et al. \cite{engstrom2019exploring} identified that even simply rotating and/or translating the benign images can result in a significant degrade of classification performance in DNNs.
Xiao et al. \cite{xiao2018spatially} proposed an adversarial attack method that can generate perceptually realistic adversarial examples by perturbing the spatial locations of pixels. 
However, those non-additive methods are only able to generate specific perturbation that is workable on a given image, rather than a universal one that can fool a deep neural network over the whole dataset. 

Therefore, in this paper, we first aim to design a novel universal adversarial attack method that can generate non-additive perturbation (i.e., spatial transformation) to fool DNNs over a large number of inputs simultaneously. We then try to further surpass current universal attack approaches with an aim to unify both additive and non-additive perturbations under a same universal attack model.
As a result, we propose a unified and flexible framework, called GUAP, that can capture the distribution of the unknown additive and non-additive adversarial perturbations jointly for crafting universal adversarial examples. Specifically, the generalized universal adversarial attack can be achieved via spatial transformed or $\ell_\infty$-norm based perturbations or the combination of both. 
In summary, the novelty of this paper lies in the following aspects:
\begin{itemize}[leftmargin=*]
    \item We propose a novel unified framework, GUAP, for universal adversarial attacks. 
    As the first of its kind, GUAP can generate either additive (i.e., $\ell_\infty$-bounded) or non-additive (i.e., spatial transformation) perturbations, or a combination of both, which considerably generalizes the attacking capability of current universal attack methods.
    \item As far as we know, GUAP is also the very first work that can initiate universal adversarial attacks on DNNs by spatial transformations. We show that, with the help of spatial transformation, GUAP is able to generate less distinguishable adversarial examples with significantly better attacking performance than existing state-of-the-art approaches, leading to over $15\%$ and $19\%$ improvements on various CIFAR-10 and ImageNet DNNs in terms of attack success rate.
    \item The proposed method fits the setting of semi-white attack, which can synthesize adversarial images without accessing the structures and parameters original target model.
    In addition, with the universal and input-agnostic property, the produced perturbation can be used directly in the attacking phase without any further computation, which provides excellent efficiency in practice.
\end{itemize}

\section{Related Work}
\label{related}

We first discuss the related works in universal adversarial attacks, which are all based on additive perturbations so far. Then we review spatial-based adversarial attacks which are all local and non-universal. As far as we know, there is no existing work that can exactly achieve same functionalities as ours. Table~\ref{methods} indicates such uniqueness of our research. 

\subsection{Universal Adversarial Attacks}

UAP proposed by Moosavi-Dezfooli et al.~\cite{moosavi2017universal} is the very first work that identifies the vulnerability of DNNs to universal adversarial perturbations. To create a universal perturbation, UAP integrates the learned perturbations from each iteration. If the combination cannot mislead the target model, UAP will find a new perturbation followed by projecting the new perturbation onto the $\ell_p$ norm ball to ensure it is small enough and meets the distance constrain.
This method will keep running until the empirical error of the sample set is sufficiently large or a threshold error rate is satisfied.
The optimization strategy for seeking the minimal noise is ported from `DeepFool'~\cite{moosavi2016deepfool},  a popular $\ell_2$ adversarial attack.

There are three previous works leveraging the generative model for universal adversarial attacks~\cite{hayes2018learning,poursaeed2018generative,reddy2018nag}.
All of them attempt to capture the distribution of the adversarial perturbation, which show some improvements than UAP~\cite{moosavi2017universal}.
However, their implementations have some differences.
`Universal Adversarial Network' (`UAN')~\cite{hayes2018learning} is composed of stacks of deconvolution layers, batch normalisation layers with activation function, and several fully-connected layers on the top.
For UAN, it includes a $\ell_p$ distance minimization term in the objective function, and the magnitude of the generated noise is controlled by a scaling factor, which increases gradually during training.
Poursaeed et al.~\cite{poursaeed2018generative} employed a ResNet generator introduced in~\cite{johnson2016perceptual} for generating universal adversarial perturbation, named `GAP'.
Before adding the perturbation to an image, the constraint of the noise is restricted by a fixed scaling factor directly.
In addition, this work was also extended for semantic segmentation task.
Mopuri et al.~\cite{reddy2018nag} proposed a generative model called `NAG', which consists 5 deconvolution layers and a fully-connected layer.
NAG presented an objective function which aims to reduce the benign prediction confidence and increase that of other categories.
Besides, a diversity term was introduced within the objective function to encourage the diversity of perturbations.

Furthermore, Mopuri et al. \cite{mopuri2017fast,mopuri2019generalizable} had illustrated that it is possible to craft a single perturbation that can fool most of the images under a data-free setting by maximizing the output after activation function over multiple layers in the target model, but this may compromise the attack success rate compared to UAPs~\cite{moosavi2017universal}. 

\begin{table}[!h]
\caption{Comparison among the Existing Related Work}
\centering
\renewcommand\arraystretch{1.5}
\begin{tabular}{c|c|c|c}
\hline
\hline
\multirow{2}{*}{Method} & \multicolumn{3}{c}{Attacking Capability} \\ \cline{2-4}
 & additive & non-additive & universal \\ \hline\hline
UAP~\cite{moosavi2017universal} & $\surd$ &  & $\surd$ \\
FFF~\cite{mopuri2017fast} & $\surd$ &  & $\surd$ \\
UAN~\cite{hayes2018learning} & $\surd$ &  & $\surd$ \\
GAP~\cite{poursaeed2018generative} & $\surd$ &  & $\surd$ \\
NAG~\cite{reddy2018nag} & $\surd$ &  & $\surd$ \\
StAdv~\cite{xiao2018spatially} &  & $\surd$ &  \\
Engstrom et al.~\cite{engstrom2019exploring} &$\surd$ &$\surd$&\\
GUAP (Ours) & $\surd$ & $\surd$ & $\surd$ \\ \hline\hline
\end{tabular}%
\label{methods}
\end{table}

\subsection{Spatial-based Adversarial Attacks}

Fawzi \& Frossard~\cite{fawzi2015manitest} firstly studied the in-variance of deep networks to spatial transformations, which suggests convolutional neural networks are not robust through rotations, translations, and dilation.
Xiao et al. \cite{xiao2018spatially} also argued that the traditional $\ell_p$ constraint may not be an ideal criterion for measuring the similarity between two images. 
They proposed a spatially transformed optimization method, which is capable of generating perceptually realistic adversarial examples with high fooling rate by changing the positions of pixels, rather than adding perturbation to the clean image directly.
It manipulates an image according to a pixel replacement rule named ‘flow field’, which maps each pixel in the image into a new value. 
To ensure that an adversarial image is perceptually close to the benign one, they also minimize the local geometric distortion instead of the $\ell_p$ distance in the objective function.
Xiao et al.~\cite{xiao2018spatially} also conducted a human perceptual study, which proved that the spatially transformed adversarial perturbations are more indistinguishable for humans, compared with those additive perturbations generated by \cite{goodfellow2014explaining,carlini2017towards}.

Engstrom et al. \cite{engstrom2019exploring} noted that 
existing adversarial methods are too complicated, and generate contrived adversarial examples that are highly unlikely to occur `naturally'. 
They thus showed neural network is even quite vulnerable to simple rotations.
They constrained the transformation in a range of $\pm30^\circ \times \pm3$ pixels, then by conducting grid search, which explores the parameters space and exhaustively testing all possibilities, simply rotating and/or translating the benign images can result in the significant degrade of the target classifier’s performance.
Furthermore, a combination adversary is also be considered, which performs all possible spatial transformations (through exhaustive grid search) and then applies a $\ell_p$-bounded PGD attack \cite{madry2017towards} on top.
From the experimental observation, they further indicated that the robustness of these two kinds of perturbation are orthogonal to each other. 

As Table~\ref{methods} shows, our approach, GUAP, is much more advanced than all current adversarial attacks in terms of attacking capability which is not only universal but also could be additive, non-additive or both.

\begin{figure*}[t]
\centering
\includegraphics[width=\textwidth]{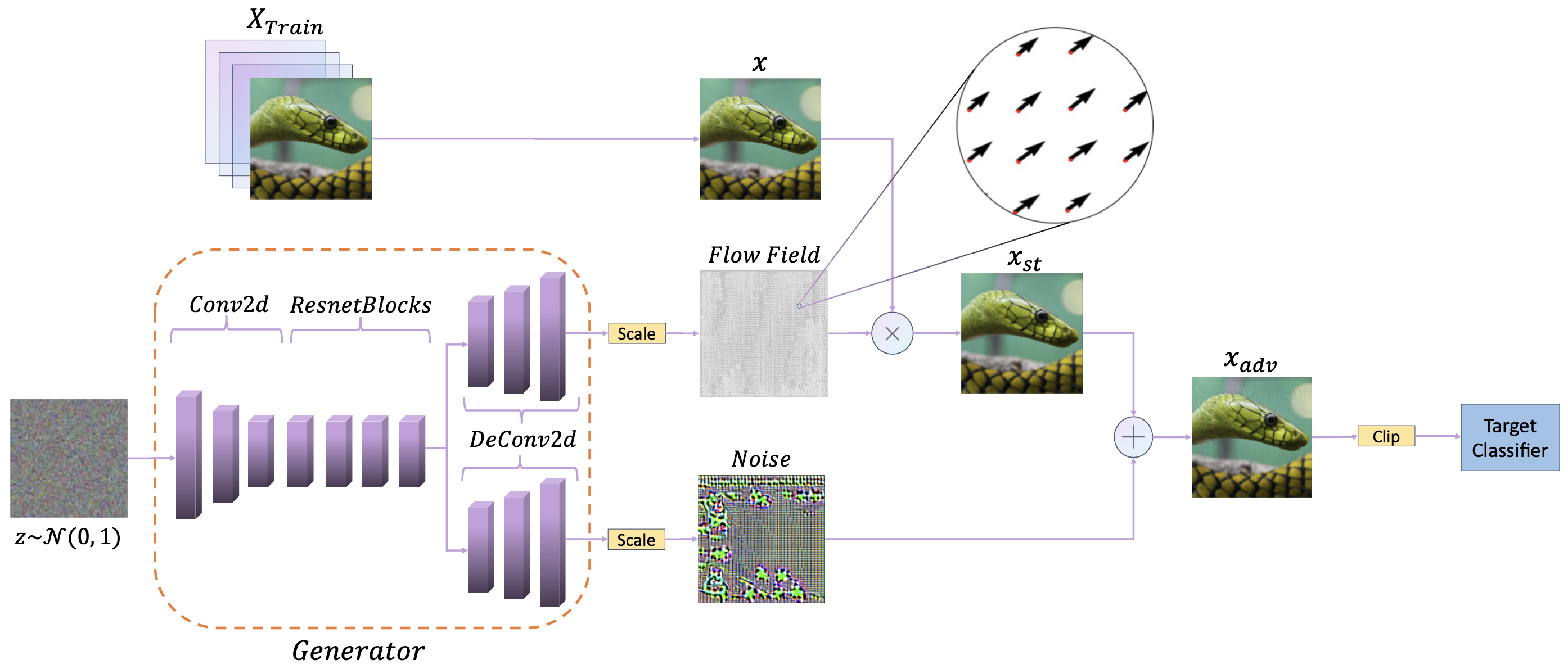}
\caption{Overview of Generalized Universal Adversarial Perturbation, here $\otimes$ represents the spatial transformation operation, $\oplus$ is the additive implementation.} 
\label{process}
\end{figure*}

\section{Generalized Universal Adversarial Perturbations}
\label{method}

Previous research on UAPs is based on $\ell_p$-bounded attack, which requires generated images to be close to the benign examples within a given $\ell_p$ norm ball.
This is based on the assumption that human perception can be quantified by $\ell_p$ norm.
However, for a non-additive method like spatial transformation, $\ell_p$ norm is difficult to capture the perceptional indistinguishability of humans, as proved in~\cite{engstrom2019exploring}.
On the other hand, prior non-additive approaches only generate specific perturbations for a given image, rather than a universal one over the whole dataset.

As a result, we propose a framework to work with both the additive ($\ell_p$ norm based) and non-additive (spatial transformation  based) methods to craft the \underline{G}eneralized \underline{U}niversal \underline{A}dversarial \underline{P}erturbations (GUAP).
The proposed framework leverages the training process in an end-to-end generative model for generating a universal flow and a universal noise jointly.
Figure~\ref{process} indicates the whole workflow for the generation of universal perturbations. We will detail our framework in this section.

\subsection{Problem Definition}

Given a benign data set $\mathcal{X} \in \mathbb{R}$ from $\mathcal{C}$ different classes, and a target DNN classifier $h: \mathcal{X} \to \mathcal{Y}$, which assigns a label $h(x)\in\{1,2,...,\mathcal{C}\}$ to each input image $x$.
We assume that all features of the images are normalized to range $[0,1]$, and $h$ has achieved high accuracy on the benign image set. 
Moreover, we denote $\mathcal{A}$ as the space of adversarial examples, such that the corresponding adversarial example $x_{adv} \in \mathcal{A}$ is able to fool target model $h$ with high probability while resembles the natural image $x$. 
For untarget adversarial attacks, we can express this as $h(x) \ne h(x_{adv})$ formally, meanwhile satisfies the defined distance metric.
Traditional universal attacks focus on finding a universal noise $\delta$ for all inputs, which generates adversarial examples $x+\delta$ that are able to fool the target classifier. 
The maximum perturbation constraint is controlled by $\|\delta\|_\infty \le \epsilon$.

However, in our case, we will consider both non-additive and additive perturbations, i.e. spatial based and $\ell_\infty$ norm based.
The adversarial sample $x_{adv}$ with respect to \emph{any} input data $x$ can be represented as:
\begin{equation}
    x_{adv} = \mathcal{F}_f(x)+\delta
\end{equation}
where $\mathcal{F}_f(\cdot)$ is the adversarial flow function for spatial transformation~\cite{xiao2018spatially}.
$f$ and $\delta$ are the trainable \emph{universal} flow field and \emph{universal} noise for performing spatial based perturbation and $\ell_\infty$-bounded perturbation, respectively, over the whole dataset.
We inherit a hyper-parameter $\epsilon$ from traditional UAPs to control the magnitude of the universal noise and further introduce another parameter $\tau$ to limit the perturbation caused by the spatial transformation.

\subsection{Universal Spatial Transformations}
Spatial transformed attack was proposed in~\cite{xiao2018spatially}.
It is an image-specific method that optimizes flow fields for different input images, and generate adversarial example by manipulating each pixel value based on a learned flow field.
We also utilize the flow field to capture the spatial transformation but in an image-agnostic manner.

Formally, we define the image input space $X \in [0,1]^{c \times h \times w}$. 
A universal flow field $f \in [0,1]^{2 \times h \times w}$ represents a pixel replacement rule:
for a pixel $x^{(i)}$ at the location ($u^{(i)}$,$v^{(i)}$), its corresponding coordinate in $f$, i.e., 
$f_i = $ ($\Delta u^{(i)}$, $\Delta v^{(i)}$),  denotes the direction and magnitude for the replacement of pixel value of $x^{(i)}$.
Let $x_{st}$ stands for the spatial transformed image from the benign image $x$ via the flow field $f$, the relation between the renewed coordinate and the original pixel location can be expressed as: 
\begin{equation}
\big(u^{(i)}, v^{(i)}\big)=\big(u_{st}^{(i)}+\Delta u^{(i)}, v_{st}^{(i)}+\Delta v^{(i)}\big), 
\end{equation}
for all $i $ in $\{1, \ldots, h \times w\}$. 
Note that, in this paper, the same flow field ($\Delta u^{(i)}$, $\Delta v^{(i)}$) is applied to all the channels for a given pixel.
Since a pixel coordinate only accepts integer format, the flow field with shape $(2 \times h \times w)$ is necessary for handling pixel transformation, which allows the flow field to transform a pixel value to a location along the vertical and horizontal directions respectively even it is not lying on the integer-grid.
To make sure that $f$ is differentiable during training, bi-linear interpolation~\cite{jaderberg2015spatial} is employed to compute an appropriate pixel value over the immediate neighbourhood for the transformed image $x_{st}$:
\begin{equation}
    x_{st}^{(i)}=\sum_{q \in N\big(u^{(i)}, v^{(i)}\big)} x^{(q)}\big(1-\big|u^{(i)}-u^{(q)}\big|\big)\big(1-\big|v^{(i)}-v^{(q)}\big|\big)
\end{equation}
Here the neighbourhood $N(\cdot)$ is the defined four positions of (top-left, top-right, bottom-right, bottom-left) tightly surrounding the targeted pixel. 
Such that the spatially perturbed image remains the same shape as the original image.
To encouraged the flow field $f$ to be \emph{small} enough and generate images with high perceptual quality, the flow loss based on total variation~\cite{rudin1992nonlinear} is introduced for enforcing the local smoothness:
\begin{equation}
\begin{aligned}
    &\mathcal{L}_{flow}(f) =\\
    &\sum_{p}^{\textrm{all pixels}} \sum_{q \in \mathcal{N}(p)}\sqrt{||\Delta u^{(p)}-\Delta u^{(q)}||^2_2 + ||\Delta v^{(p)}-\Delta v^{(q)}||^2_2}
\end{aligned}
\end{equation}
This flow loss is included in the objective function, together with a fooling loss introduced in \cite{carlini2017towards}, which will be minimized during training.
However, the hyper-parameter for balancing these two losses in each dataset may different, thus it becomes unclear when measuring the magnitude of the spatial distortion.

\begin{figure}[!h]
\centering
\includegraphics[width=0.4\columnwidth]{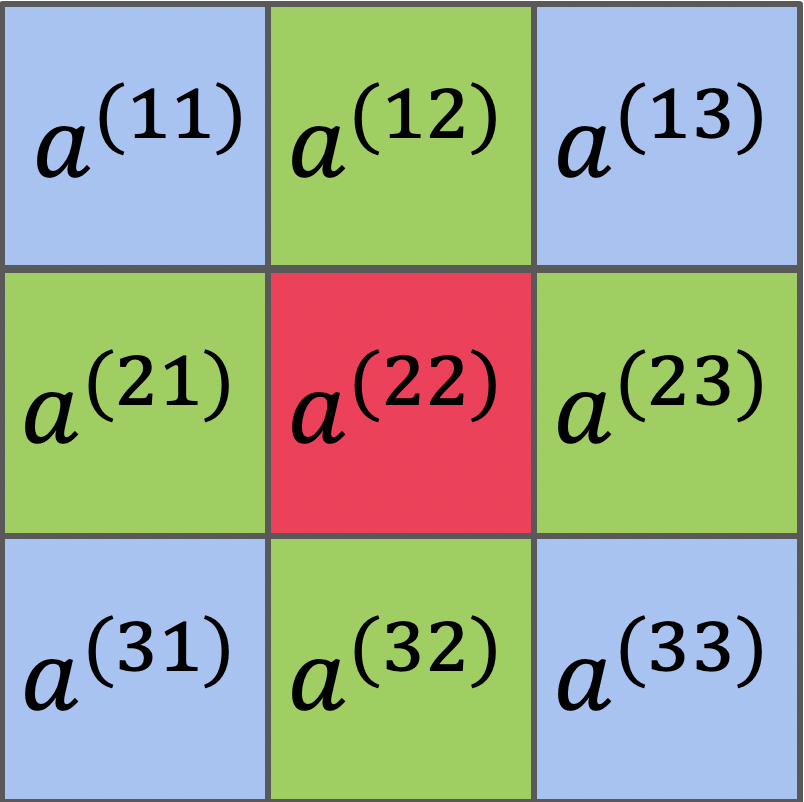}
\caption{The Von Neumann neighbours of red pixel $a^{(22)}$ include four green pixels $a^{(12)}$, $a^{(21)}$, $a^{(23)}$ and $a^{(32)}$.} 
\label{neighbour}
\end{figure}

When applying bi-linear interpolation, the new value of $x_{st}$ depends on the direction and magnitude that the original $x$ changes towards.
As we want the perturbation to be as imperceptible as possible, intuitively we can constrain that each single pixel can only move its mass to nearby neighbors.
Different from~\cite{xiao2018spatially}, here we introduce a hyper-parameter $\tau$ to budget the perturbations caused by the spatial perturbation, which satisfies the constraint: $L_{flow}(f) \le \tau$, here the $L_{flow}(f)$ is defined as:
\begin{equation}
\begin{small}
\begin{aligned}
&L_{flow}(f) =\\
&\max_{q_j \in \mathbb{N}(p)}\left( \sqrt{\frac{1}{n}\sum_{p}^{n} \|\Delta u^{(p)}-\Delta u^{(q_j)}\|^2_2 + \|\Delta v^{(p)}-\Delta v^{(q_j)}\|^2_2}\quad\right)
\end{aligned}
\end{small}
\end{equation}
Here $n$ is the number of pixel for the corresponding image, and $\mathbb{N}$ is the function of the defined Von Neumann neighbourhood~\cite{toffoli1987cellular}.
For an image, Von Neumann neighbourhood defines the notion of 4-connected pixels which have a Manhattan distance of 1, as shown in Table~\ref{neighbour}. 
This constraint can be further derived as:
\begin{equation}
\label{eq}
    \|\Delta u^{(p)}-\Delta u^{(q)}\|^2_2+\|\Delta v^{(p)}-\Delta v^{(q)}\|^2_2 \le \tau^2
\end{equation}
Intuitively speaking, for each pixel, this implies that the explicit $\tau$ budgets the mass it can move along the horizontal and vertical direction.
This notion is a critical component for quantifying the intensity of the spatial transformation, rather than optimizing a loss with arbitrary value in the objective function.
If we assign the value of $\tau$ as $0.1$, and consider an extreme case when one of the terms on the left side in (\ref{eq}) is zero, it will merely moves towards one direction (horizontal or vertical) with 10\% of the pixel mass by one pixel.
Intuitively speaking, this can also be understood as moving less than 10\% of the pixel mass along two orientations for more than one pixel. 

This concept can be approximately regarded as the notion of Wasserstein distance introduced in \cite{wong2019wasserstein}, which projects the generated noise onto a Wasserstein ball by employing a low-cost transport plan.
We also restrict the spatial transformation in a local configuration, which only moves the adjacent pixels mass to the nearby pixels.
The key difference is that Wong et al.~\cite{wong2019wasserstein} utilized the sinkhorn iterations for calculating the projected Wasserstein distance relying on the additive noise, while in our method, this is achieved via interpolating from its neighbourhood without extra additive perturbation, which is more natural for image manipulations and leads to a quasi-imperceptible effect for human's eye.

\begin{figure}[!h]
\centering
\includegraphics[width=\columnwidth]{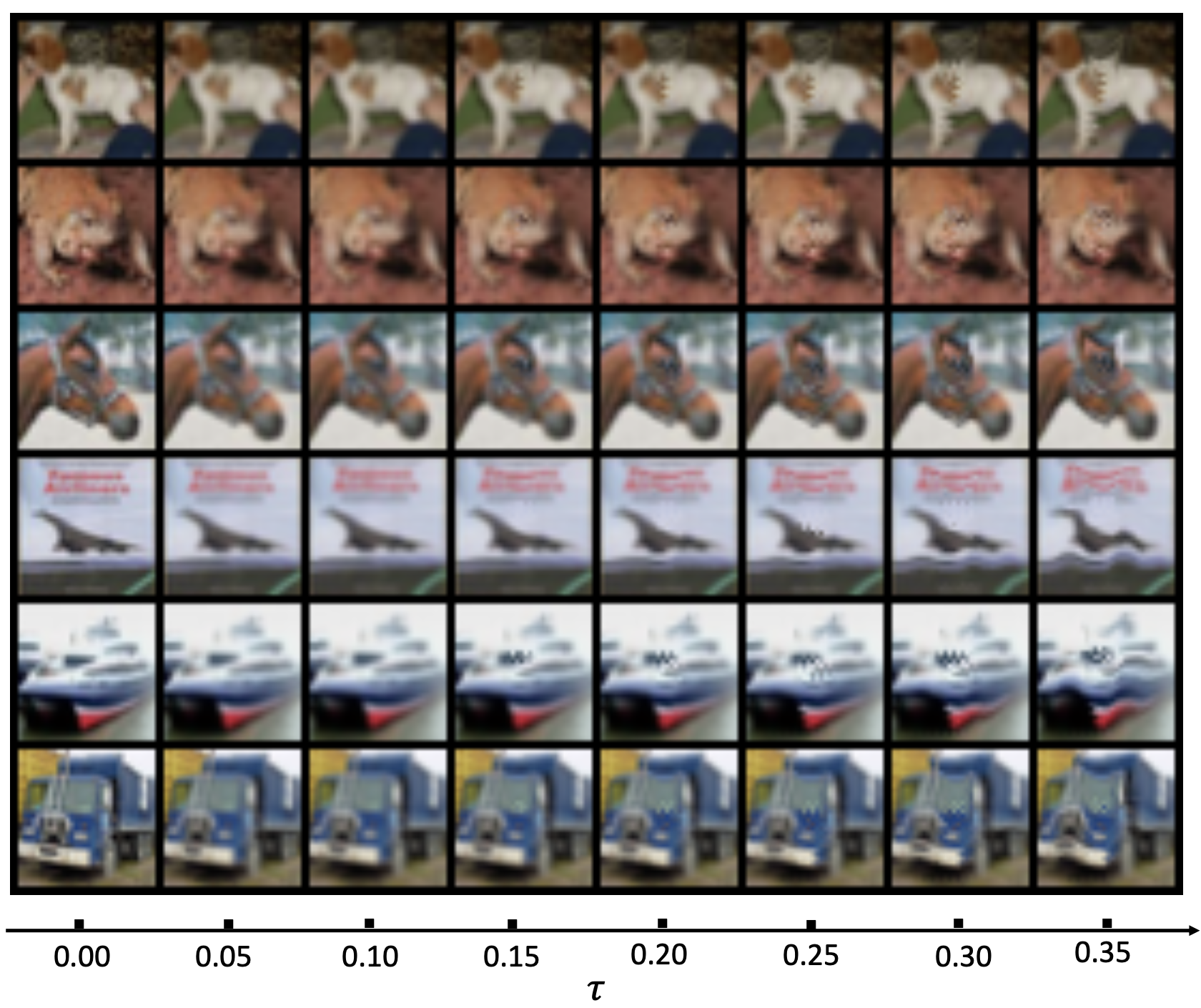}
\caption{Spatially transformed adversarial images with the increase of $\tau$.} 
\label{tau}
\end{figure}

It is noted that different images have different reactions on spatial perturbation.
For example, as shown in Fig.~\ref{tau}, the deformations in the first three rows are almost indistinguishable for human beings even with large $\tau$, however, in the last three rows, the distortion is perceptible when $\tau$ is larger than 0.1.
For those images who have the regular structure like straight line, their distortion leads to a curve shape and becomes noticeable for human's eyes.
In natural creation, most things are not in a straight line shape, which makes this spatial perturbation useful in the physical world.
In empirical, the setting $\tau=0.1$ with indistinguishable visual quality is adapted for our experiments in Section~\ref{experiment}, where we will show how it can help for universal perturbation.

\subsection{Generalized Adversarial Generator}
In our unified framework, there are two major components for producing the generalized universal adversarial perturbations: {\em 1)} universal spatial perturbation; {\em 2)} universal noise perturbations. 
If we set $\tau$ to 0, the spatial transformation will turn to identity transformation, and our framework will reduce it to craft universal additive perturbation only.
On the other hand, when $\epsilon$ is set to 0, our framework results in learning spatially transformed universal perturbations.
Most importantly, our framework enables it to work collaboratively and generalize universal adversarial attacks by considering both additive and non-additive perturbations.

Inherited the benefits from \cite{hayes2018learning,poursaeed2018generative,reddy2018nag}, we employ the generative model for seeking a small universal noise, rather than the iterative approach in~\cite{moosavi2017universal}.
Differently, we also capture the distribution of the universal spatial perturbation at the same time.
By doing so, the learned perturbations will not directly depend on any input image from the dataset, so called universal perturbations.
Here we show how to learn these two universal perturbations jointly
via an end-to-end generative model.
We adapt the similar architectures as image-to-image translation adversarial networks~\cite{zhu2017unpaired,isola2017image,xiao2018generating} for perturbation generation, which utilizes an encoder-bottle\_neck-decoder structure to transfer an input pattern to the desired outputs. 
The overall structure of the generator can be seen in Fig.~\ref{process}, the encoder is consist of 3 Convolution layers followed by instance normalization and ReLU.
After going through 2 ResNet blocks, there are two decoders producing two outputs for learning different universal perturbations, respectively, each of them contains 3 Deconvolution layers, followed by instance normalization and ReLU activation functions again.

The largest difference to other generative models for universal learning~\cite{hayes2018learning,poursaeed2018generative,reddy2018nag} is that we only take a single input noise vector as input, i.e. batch-size equals 1, meanwhile apply instance normalization during training. 
The outputs of generator consist of two parts: a universal flow field $f$, and a small universal perturbation noise $\delta$, thus we apply another decoder for crafting spatial perturbation.
In addition, we eschew the discriminator in generative adversarial networks (GAN)~\cite{goodfellow2014generative} because the natural-looking attribute has already been controlled by the $\epsilon$ and $\tau$, respectively.

Formally, for our generalized adversarial generator $G_\theta(z)$ parameterized by $\theta$, it is fed a random pattern $z\sim \mathcal{N}(0,1)$ and outputs a flow field $f_0 \in [0,1]^{2\times h \times w}$ and a noise $\delta_0 \in [-1,1]^{c\times h\times w}$,
activated by a \emph{Sigmoid} function and a \emph{Tanh} function, respectively.
Then the output $f_0$ is scaled to obtain the universal flow field $f$. This operation ensures the prerequisite constraint for spatial distortion is met, controlled by parameter $\tau$:
\begin{equation}
    f= \frac{\tau}{L_{flow}(f_0)}
\end{equation}
which is then applied to any $x \in X$ to perform spatial distortion.
On the other hand, in terms of the output noise, we also scale it to satisfy the $\ell_\infty$ constraint, like assigning $\epsilon=0.03$
for color image values ranged from $[0, 1]$:
\begin{equation}
    \delta = \frac{\epsilon}{\|\delta_0\|_\infty}
\end{equation}
In the next step, we combine it with the spatially perturbed image generated by the learned flow field $f$,
then the clipping implementation is be conducted to guarantee that each pixel in the image has a valid value in $[0, 1]$.
The final adversarial example can be expressed as:
\begin{equation}
    x_{adv}=\mathcal{F}_f(x)+\delta
\end{equation}
Such that we can define a loss function that leads to misclassification and update the parameters of the generator accordingly. 
Overall, we summarize the proposed method GUAP in Algorithm~\ref{guap}.

\begin{algorithm}[!h]
 \caption{Generalized Universal Adversarial Perturbations}
 \label{guap}
 \begin{algorithmic}[1] 
    \Require Training set $X$, total epochs $T$, initial input pattern $z$, adversarial radius $\epsilon$, $\mathcal{G}_\theta$, maximum flow $\tau$ , target model $h$, and number of mini-batches~$M$
    \Ensure  A universal flow field $\mathcal{F}$ and a universal perturbation noise $\delta$
    \For{$t = 1 \ldots T$}
        \For{ $i = 1 \ldots M$}
             \State $x = X_i, z = \mathcal{N}(0,1)$ 
             \State $\delta_0,f_0 = \mathcal{G}_\theta(z)$
             \State $f = \tau/L_{flow}(f_0), \delta = \epsilon/\|\delta_0\|_\infty$
             \State $x_{adv} = \mathcal{F}_f(x)+\delta$
             \State $x_{adv} = {\rm Clip}(x_{adv}, 0, 1)$
            \State // Update model weights with loss function $L_{adv}$
            \State $\theta=\theta-\nabla_{\theta} L_{adv}\left(h(x_{adv}), h(x)\right)$ 
        \EndFor
    \EndFor
 \end{algorithmic}
\end{algorithm}

\subsection{Objective Function}

Given an input $x$, the corresponding adversarial examples $x_{adv}$ aims at fooling the target classifier $h$ with high probability.
We denote the original prediction of $x$ as $y:=\arg \max h(x)$, then the objective function for an untargeted generalized perturbation attack attempts to find the universal perturbation which leads the original prediction to a wrong class.
Here we define it as: 
\begin{equation}
L_{a d v}(x, \mathcal{F},\delta)= - \hat{l}_{ce}(h\left(x_{adv}\right),y)
\end{equation}
where $\hat{l}_{ce}$ is the surrogate loss function of the conventional cross-entropy $l_{ce}$:
\begin{equation}
\hat{l}_{ce}(h\left(x_{adv}\right),y)= \frac{1}{N}\sum_{i=1}^{N}\log( l_{ce}(h\left(x_{adv}^i\right),y)+1)
\end{equation}
Since there is no upper bound for the traditional cross-entropy loss, one single evaluating data point is potentially able to cause arbitrary loss value from $0$ to $\infty$.
The worst case happens when there is a single image turned into a perfect adversarial example, this causes misclassification, but it dominates the cross entropy loss and forces the average loss to infinity.
There is no doubt that this raises the difficulty of finding the optimal parameter and leads to slow convergence. 
To tackle this problem, we propose the scaled cross entropy above to enforce our optimizer to search for the perturbation that targets at as many as data points as possible.
The `$+1$' operation ensures the output of the log function remains positive.
On the other hand, the natural log function scales the original loss for each image. This avoids any single images standing over the objective during the optimization.  
We intensively evaluate the effectiveness of this scaled loss function in Section~\ref{experiment}.

\section{Experiments}
\label{experiment}

Extensive experiments are conducted on two benchmark image datasets to evaluate the performance of the proposed framework, i.e. CIFAR-10~\cite{krizhevsky2009learning} and ImageNet~\cite{deng2009imagenet}.
The code\footnote{Our code is available at \url{https://github.com/TrustAI/GUAP}} is created with PyTorch library. 
The experiments run on one or two GeForce RTX 2080Ti GPUs.

The proposed method can have several different setups controlled by two parameters, i.e., $\epsilon$ and $\tau$.
We refer the configuration ($\epsilon=0.04,\tau=0$) as GUAP\_v1, which only performs the universal attack under the traditional $\ell_\infty$ norm.
As regarding to the combination attack, for convenience we refer the setup, GUAP\_v2: $\epsilon=0.03, \tau=0.1$; and GUAP\_v3: $\epsilon=0.04, \tau=0.1$.
A small value of $\tau$ makes sure the caused spatial distortion is imperceptible to humans, while we show that combining with small $\ell_\infty$-bounded perturbation, can achieve state-of-the-art results on both small and large benchmark datasets.

\subsection{Universal Attack on CIFAR-10 Dataset}

\begin{table}[!h]
\centering
\caption{Comparison with the State-of-the-art Universal Attack Methods on CIFAR10 Dataset}
\renewcommand\arraystretch{1.5}
\begin{tabular}{c||c|c||c||c||c}
\hline\hline 
\multirow{2}{*}{Attack}                      & \multicolumn{2}{c||}{Configuration} & VGG19  &ResNet101 & DenseNet121   \\ \cline{2-6}
& $\epsilon$&$\tau$ &ASR & ASR&ASR
\\\hline\hline
UAP      &  0.04                               &   -                              &                                                     57.2\%                                    &                                  76.0\%    &67.9\%    \\ \hline
FFF &                 0.04               &    -                             &          20.1\%    &  36.5\%                                                            &  34.1\%                                                     \\ \hline
UAN              &   0.04                       &     -                                                  &                 66.6\%                                                &           85.1\%     & 75.0\%       \\ \hline
GUAP\_v1  &  0.04&       0.00         &  \bf{84.25\%}  &  \bf{88.58\%}  &   \bf{89.23\%}                            \\\hline 
GUAP\_v2  &  0.03&       0.10    & \bf{86.86\%}                & \bf{89.45\%} &   \bf{90.09\%} \\\hline
GUAP\_v3  &  0.04&       0.10 & \bf{89.59\%} & \bf{89.56\%}& \bf{90.09}\%\\
\hline\hline
\end{tabular}
\label{tablecifar}
\end{table}

\begin{figure*}[!h]
\centering
\includegraphics[width=0.7\textwidth]{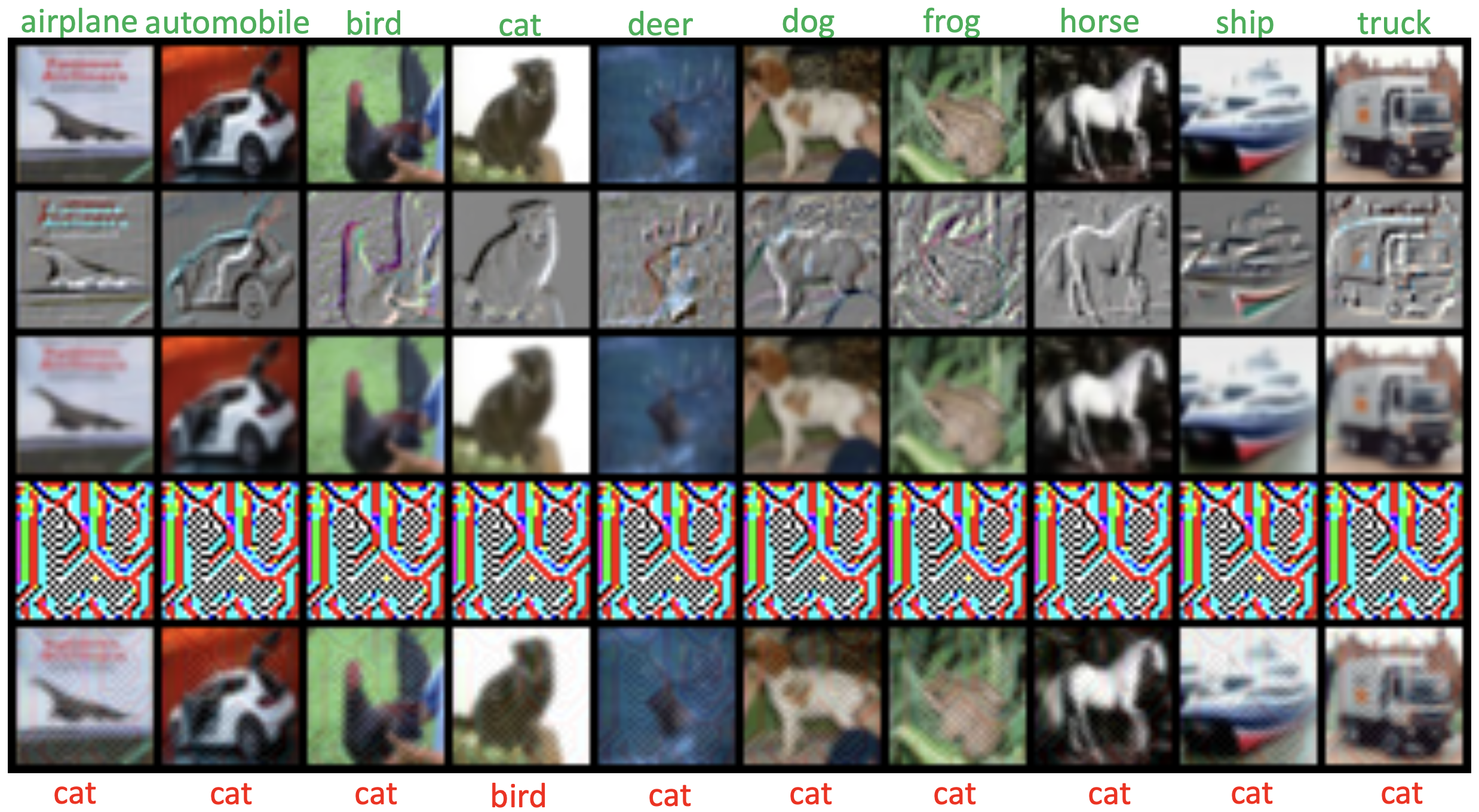}
\caption{Attacking performance of GUAP\_v2 on CIFAR-10 dataset among 10 classes against VGG19.} 
\label{cifarimage}
\end{figure*}

CIFAR-10 dataset contains 60,000 colour images from ten different classes, with 6,000 images per class.
Each image has $32\times32$ pixels. 
Normally they are split into 50,000 images for training purposes and 10000 images used for evaluation.
For comparison, we follow the same neural network structure as the target classifiers in~\cite{hayes2018learning} for generating universal adversarial perturbations, i.e., VGG19, ResNet101, and DenseNet121.
The standard accuracy of these models is 93.33\%, 94.00\%, and 94.79\%, respectively.

Table~\ref{tablecifar} reports the experimental results achieved by our methods and the fooling rate obtained by three universal attacks, i.e., UAP~\cite{moosavi2017universal}, Fast Feature Fool~\cite{mopuri2017fast}, and UAN~\cite{hayes2018learning}. 
Here the attack success rate (ASR) is used to measure the performance of the attack, which reflects the percentage of the images in a test set that can be used by the adversary to successfully fool the target neural networks.
In particular, when $\tau$ is set to 0, our framework degrades to craft universal noise constrained by $\ell_\infty$ norm, i.e., GUAP\_v1, but it still obviously outperforms other universal adversarial attacks in terms of ASR. We also find that VGG19 is the most resistant model to the universal attacks, which is in line with the observation in~\cite{hayes2018learning}. For this challenging VGG19 model, our method can still achieve 84.25\% ASR, with an nearly 18\% improvement than existing most strong universal attack.

\begin{figure*}[!h]
        \centering
        \subfigure[ ]
        { \label{cifarasr}
        \includegraphics[width=0.50\textwidth]{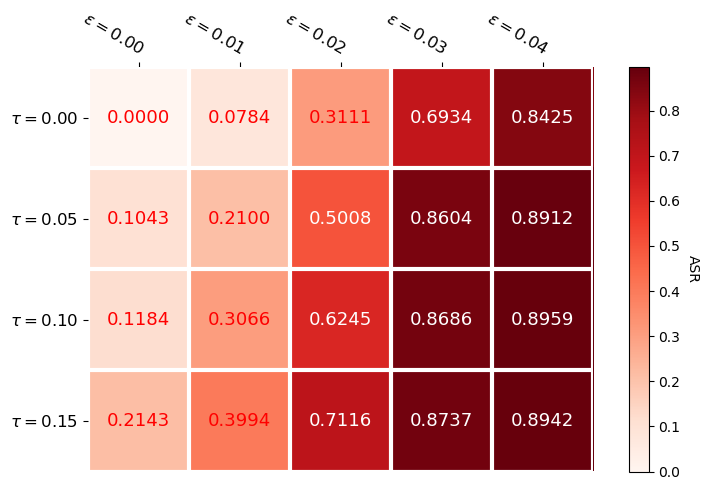}
        }
        \subfigure[ ]
        { \label{mse}
        \includegraphics[width=0.45\textwidth]{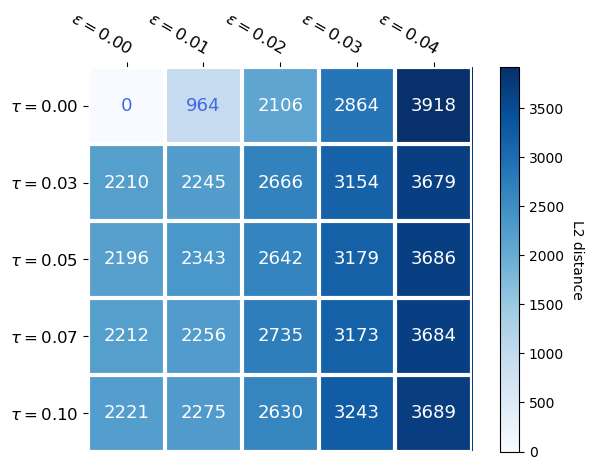}
        }
        \caption{(a) Attack success ratio for VGG19 under different combination settings on CIFAR-10 dataset;~~(b) Average $\ell_2$ distances between original images and adversarial examples for GoogLeNet on ImageNet dataset.} 
\end{figure*}

On the other hand, when $\epsilon$ equals 0, then our framework turns to universal spatial perturbation.
Here we conduct an ablation study for indicating the relationship between the spatial perturbation and the $\ell_\infty$-bounded perturbation.
The visualization in Fig.~\ref{cifarasr} demonstrates the corresponding attack success rates over the test set of the CIFAR-10, trained for the target VGG19 model.
Different heat cap colors represent the performance of the adversarial attack with different $\epsilon \in \{0.0, 0.01, 0.02, 0.03, 0.04\}$ and $\tau \in \{0.0, 0.05, 0.1, 0.15\}$.
Ahead along the x-axis, It can be observed that with the increasing magnitude of $\epsilon$, a higher fooling rate can be achieved. 
Similarly, towards the y-axis direction, a larger $\tau$ also leads to a higher attack success rate.

Since the constrained ball for spatial perturbation is different from the $\ell_\infty$ norm ball, it can be inferred that these two universal perturbations are not strictly contradictory to each other, hence performing them together brings improvements in terms of ASR but will not compromise the imperceptibility.
As expected, the proposed combination attack GUAP\_v2 is able to fool the target classifier with a very high ASR. And our attack, GUAP\_v3, obtains the highest ASR rate among all attack methods with a nearly $23\%$ improvement on VGG19 DNN.
Figure~\ref{cifarimage} displays several adversarial examples generated by GUAP\_v2 among ten different classes.
The odd rows denote the original images, spatial transformed images the final perturbed adversarial examples.
The second and fourth rows represent the differences between the original image and perturbed images caused by the spatial transform and the additive noise, respectively.
We can easily observe that the spatial-based attack mostly focuses on the edge of images, which confirms that the edge of images plays a significant role in deep neural networks.

\subsection{Universal Attack on ImageNet Dataset}
Following the instruction in~\cite{moosavi2017universal}, a subset from the training set of ILSVRC 2012 dataset~\cite{deng2009imagenet} is utilized as our training dataset, which contains 10,000 images from 1000 classes, i.e., 10 images per objects.
In addition, 50,000 images in the validation set of ImageNet are treated as the test set for evaluation.
We adapt four different neural networks as target models: VGG16~\cite{simonyan2014very}, VGG19~\cite{simonyan2014very}, ResNet152~\cite{he2016deep}, and GoogLeNet~\cite{szegedy2015going}, whose top-1 accuracy can reach 71.59\%, 72.38\%, 78.31\% and 69.78\%, respectively.
We compare GUAP with four baseline methods, including UAP~\cite{moosavi2017universal}, Fast Feature Fool~\cite{mopuri2017fast}, and two generative methods NAG~\cite{reddy2018nag} and GAP~\cite{poursaeed2018generative} for crafting universal perturbation.
 
\begin{figure}[!h]
\centering
\includegraphics[width=\columnwidth]{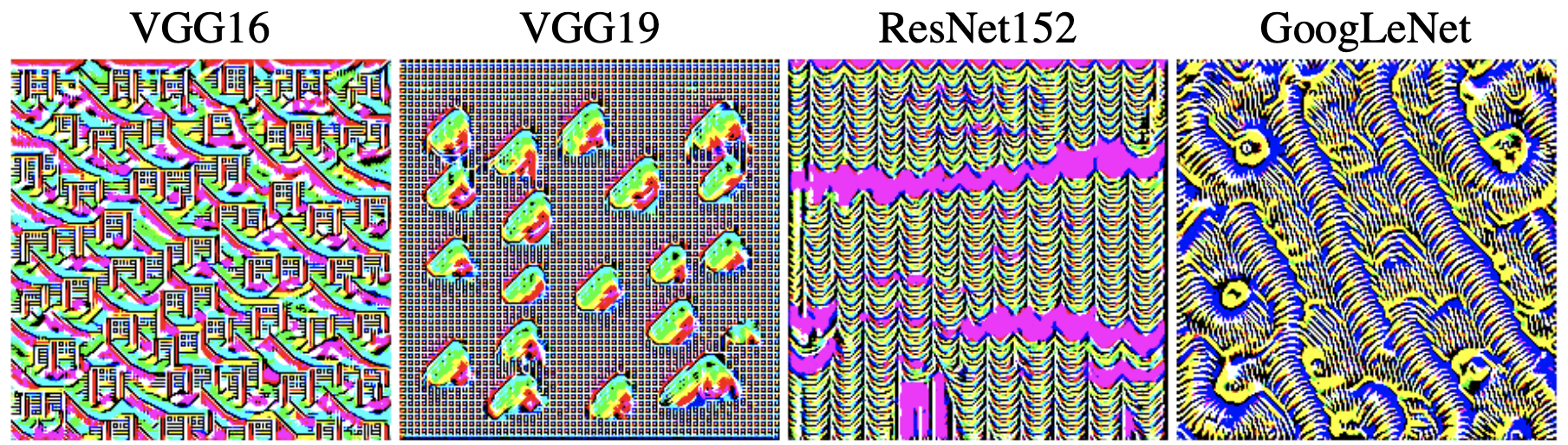}
\caption{Universal additive perturbations generated by GUAP\_v1 on ImageNet dataset.} 
\label{allnoise_imagenet}
\end{figure}

Table~\ref{imagenettable} reports the experimental results on this dataset, and several adversarial examples can be seen in Fig.~\ref{vggexample}.
We see that our proposed model GUAP\_v1 is able to obtain better fooling rates under the same magnitude of $\ell_\infty$ norm, i.e., over 90\% for all target models.
This proves the fragility of deep neural networks even the perturbation is universal.
These results exceed all other state-of-the-art methods for generating universal perturbations, which indicates an significant improvement compared to previous work, especially more than 15\% improvements on two VGG models.
UAP~\cite{moosavi2017universal} suggests that VGG19 is the most resilient DNN in ImageNet dataset.
However, we find out that GoogLeNet is the least sensitive model to additive perturbation compared to other models, so we argue that the iterative universal attack such as UAP has some limitation when applied to evaluate the robustness of DNNs. 
we visualize the learned perturbation on ImageNet for 4 target models in Figure~\ref{allnoise_imagenet}. 
We can see that perturbations from GUAP presents high level of diversities cross different models.

\begin{table*}[!h]
\centering
\caption{Comparison with the State-of-the-art Universal Attack Methods on ImageNet Dataset}
\renewcommand\arraystretch{1.5}
\begin{tabular}{c||c|c||c||c||c||c}
\hline\hline
\multirow{2}{*}{Attack Method}                      & \multicolumn{2}{c||}{Configuration} & VGG16  & VGG19& ResNet152 &GoogLeNet  \\ \cline{2-7}
& $\epsilon$&$\tau$ &Attack Success Rate &Attack Success Rate & Attack Success rate&Attack Success Rate
\\\hline\hline
UAP      &  0.04                               &   -                              &      78.30\%                                                 &                                  77.80\%    & 84.00\% &78.90\%   \\ \hline
FFF &0.04 &- &47.10\% & 43.62\% &- &56.44\%\\ \hline
NAG &                 0.04               &    -                             &          77.57\%    &  83.78\%                                                       &  87.24\%        &90.37\%                                             \\ \hline
GAP              &   0.04                       &     -                                                  &  83.70\%                      &           80.10\%     & - &82.70\%       \\ \hline
GUAP\_v1  &  0.04&       0.00             & \bf{97.06\%}&  \bf{94.80\%}           &\bf{95.15\%}  &\bf{90.60\%}                    \\\hline 
GUAP\_v2  &  0.03&       0.10    & \bf{98.25\%} & \bf{97.00\%} &\bf{96.74\%} &\bf{94.42\%}\\\hline
GUAP\_v3  &  0.04&       0.10    &\bf{98.47\%} &\bf{99.24\%} &\bf{99.03\%}& \bf{97.82\%} \\\hline\hline
\end{tabular}
\label{imagenettable}
\end{table*}

The ablation study in Fig.~\ref{compimagenet} demonstrates the interplay between the spatial perturbation and the additive noise perturbation.
When the universal $\ell_\infty$ attack is integrated with spatial perturbation, the strong attack ability can be achieved clearly.
In particular, our combination attack, GUAP\_v3, it obtains the highest attack compared to other universal $\ell_\infty$ attack using larger $\epsilon$ merely, including the proposed GUAP\_v1, showing the superior performance when the universal attack is with both additive and non-additive perturbations.
In the next section, we show that involving spatial perturbation will benefit the attacker, not only achieving the higher fooling rate but also on other properties such as increasing the human imperceptibility, reducing the training samples and improving the transferability of adversarial attack.

\section{Impacts of Spatial Perturbation}

In this section, we further investigate various properties in GUAP, especially when the spatial perturbation is involved.

\subsection{Imperceptibility of Adversarial Perturbations}
When the spatial transform is involved for adversarial attacks, even it is invisible for humans, and the learned perturbations are not strictly in the $\ell_\infty$ norm ball.
In other words, for the proposed GUAP\_v2, the constrained ball for combination attack is beyond that of the additive $\ell_\infty$ perturbations. Here, for the fairness, we use another distance metric (i.e., different to spatial displacement and $\ell_\infty$-norm), $\ell_2$-norm, to measure the difference between the original and adversarial images.
Fig.~\ref{mse} indicates the $\ell_2$ distances of the differences between the benign and the adversarial images caused by different combinations in GUAP on the ImageNet dataset.
Note that here the values for measuring similarity are calculated as an 8-bit integer valid image, giving a range of possible values from 0 to 255, which is in line with the setting in~\cite{moosavi2017universal}.

\begin{figure}[!h]
\centering
\includegraphics[width=\columnwidth]{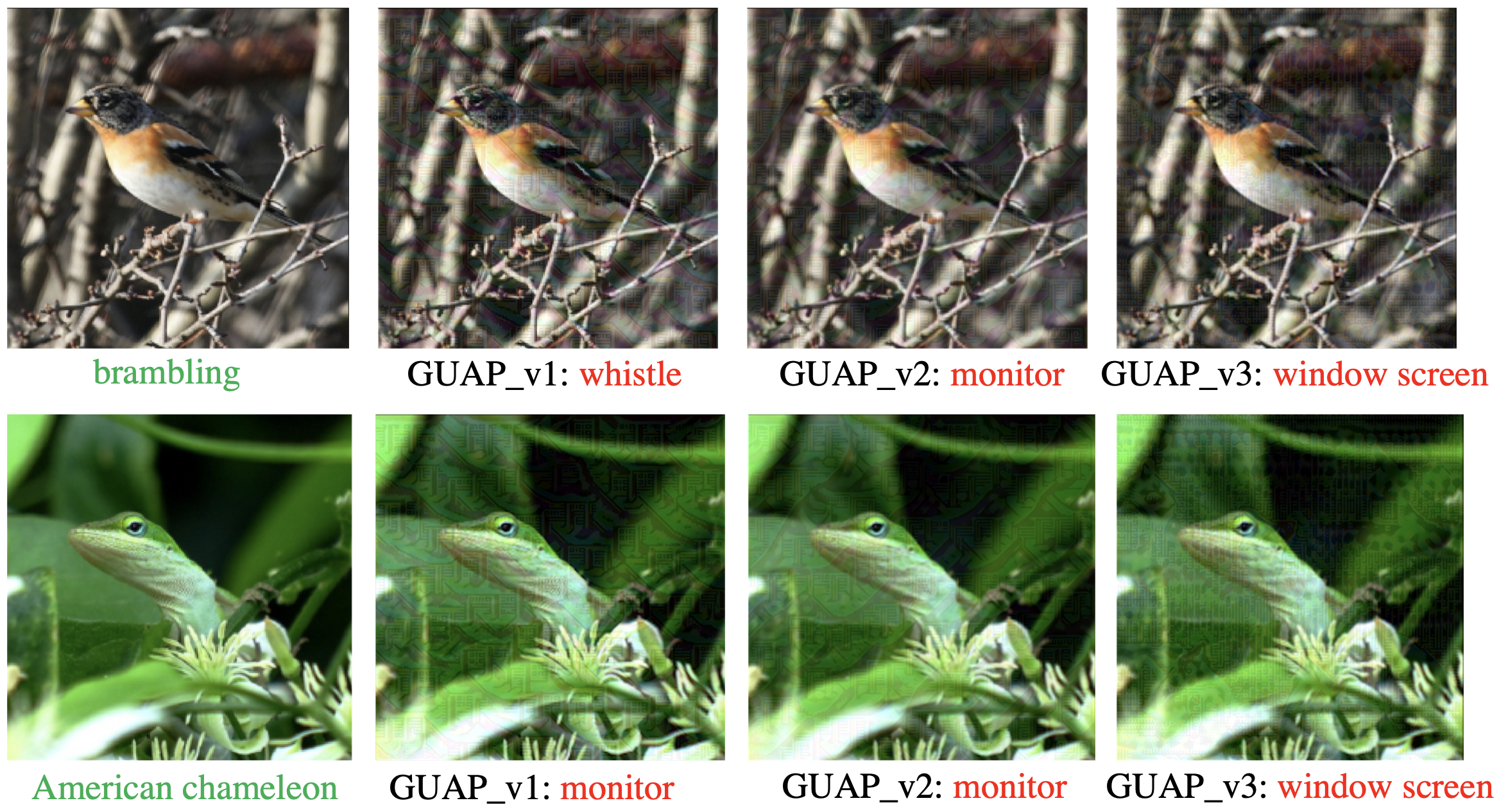}
\caption{Set of benign and corresponding adversarial examples against VGG16 on ImageNet dataset.} 
\label{vggexample}
\end{figure}

It is observed that when $\epsilon$ is larger than $0.02$, the similarity achieved by the combination attack is better than using a pure $\ell_\infty$-bounded attack with large $\epsilon$.
Since the non-additive perturbation ( i.e., spatial transform) does not modify the pixel image directly, it leads to more natural perturbations for human beings.
Thus, by combining spatially transformed perturbation with universal $\ell_\infty$ attack together, the proposed GUAP\_v2 is able to achieve a comparable or even better attack performance while has smaller $\ell_2$ distance, compared to only using the $\ell_\infty$-norm additive perturbation that usually requires larger $\epsilon$, as shown in Fig.~\ref{vggexample}.
In other words, non-additive perturbation will significantly benefit to crafting strong universal adversarial examples with better visual quality, and our method exactly provides a fixable framework to achieve this purpose.

\begin{figure*}[!h]
        \centering
        \subfigure[ ]
        { \label{compimagenet}
        \includegraphics[width=0.47\textwidth]{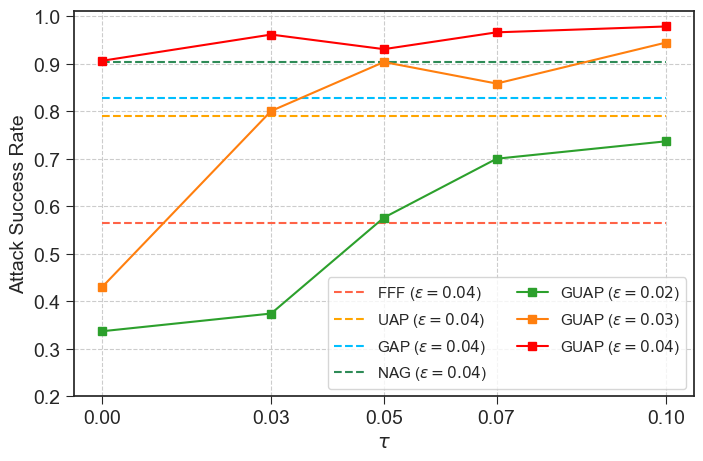}
        }
        \subfigure[ ]
        { \label{bar}
        \includegraphics[width=0.48\textwidth]{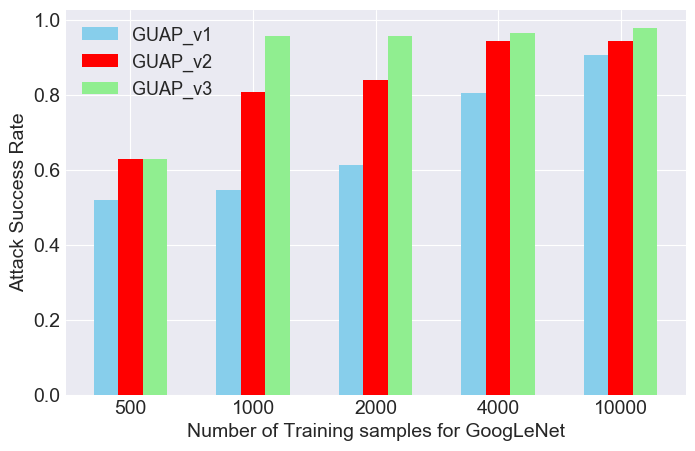}
        }
        \caption{(a) Attack success rates for VGG19 under different combination settings for GoogLeNet on ImageNet;~~(b) Attack success ratio on the validation set versus the amount of training samples.} 
\end{figure*}

\subsection{How Many Training Samples is Needed?}

There is no doubt that the number of training samples plays a crucial role not only in classification but also in adversarial attacks.  
In this section, we aim to probe the impact of the amount of training data when the spatial perturbation is taken into account for generating universal perturbation.
Here we pick up the toughest model GoogLeNet and probe the impact of the involvement of universal spatial perturbation to the attack strength.
We use four sets of training data with different sizes and generate adversarial examples on the toughest model GoogLeNet, successively.
For each setup, the proposed GUAP\_v1, GUAP\_v2 and GUAP\_v3 are conducted to search for the universal perturbations respectively, then still report the attack success rate with respect to the prediction of the target model on the whole validation set (50,000 images).

As shown in Fig.~\ref{bar}, it can be observed that with the increase in the quantity of the training samples, the fooling rate can also be improved.
Surprisingly, the combined attacking method GUAP\_v2 shows powerful capacity even though there are only 500 images used for crafting adversarial examples, which can still fool more than 60\% of the images on the validation set, which is twice more than the iterative UAP method~\cite{moosavi2017universal}, showing that
our method has a more remarkable generalization power over unseen data.
Compared with the universal $\ell_\infty$-bounded attack with 4000 training data, our combined attacking method GUAP\_v2 can use fewer images (1000) but achieve the attack success rate (80\%) at the same level. 
In particular, only 1 image per class is enough for GUAP\_v3 to produce more than 95\% fooling rate.
In other words, the combination attack GUAP\_v2/GUAP\_v3 indicate their superior capacity compared to the pure universal $\ell_\infty$-bounded attack, i.e., GUAP\_v1,
especially when only limited training samples are available.

\subsection{Transfer-based Attack}

We further explore the generalization ability of the learned UAPs cross different models.
We create 10,000 adversarial examples over the test dataset by our proposed GUAP\_v2 method, then feed them to a target classifier which was not used for learning universal adversarial examples.
As shown in Table~\ref{transfer}, our GUAP\_v2 obtains high property of transferability. 
For example, the learned UAPs on the most resilient GoogLeNet is able to mislead all other models more than 82\% fooling rate, which is very close to that obtained by training for the original classifier.
Surprisingly, the learned universal perturbation can achieve the fooling rate for these four different target models up to 90.95\% on average.
This also reveals that, with the help of spatial transformation, the transferability of learned perturbation becomes stronger.

\begin{table}[!h]\small
\centering
\caption{Transferability for GUAP\_v2 cross different models, universal perturbations are constructed using row models and tested against pre-trained column classifiers.}
\renewcommand\arraystretch{1.5}
\begin{tabular}{c|cccc}
\hline\hline
\multicolumn{1}{c|}{Model} & VGG16 & VGG19 & ResNet152 & GoogLeNet \\ \hline
\multicolumn{1}{c|}{VGG16} &        98.25\% & 90.09\% &92.62\%&93.76\%\\
VGG19                  & 93.95\%  &97.00\% &92.77\%    & 93.04\%\\
ResNet152                  &  34.75\% &31.97\% &96.74\% & 82.19\%    \\
GoogLeNet                  &  41.59\% & 31.68\%&57.95\%&94.42\%  \\
\hline
Average &67.14\% &  62.69\%&  \bf{85.02\%}  & \bf{90.85\%}\\
\hline\hline
\end{tabular}
\label{transfer}
\end{table}

\begin{figure*}[!h]
\centering
\includegraphics[width=0.96\textwidth]{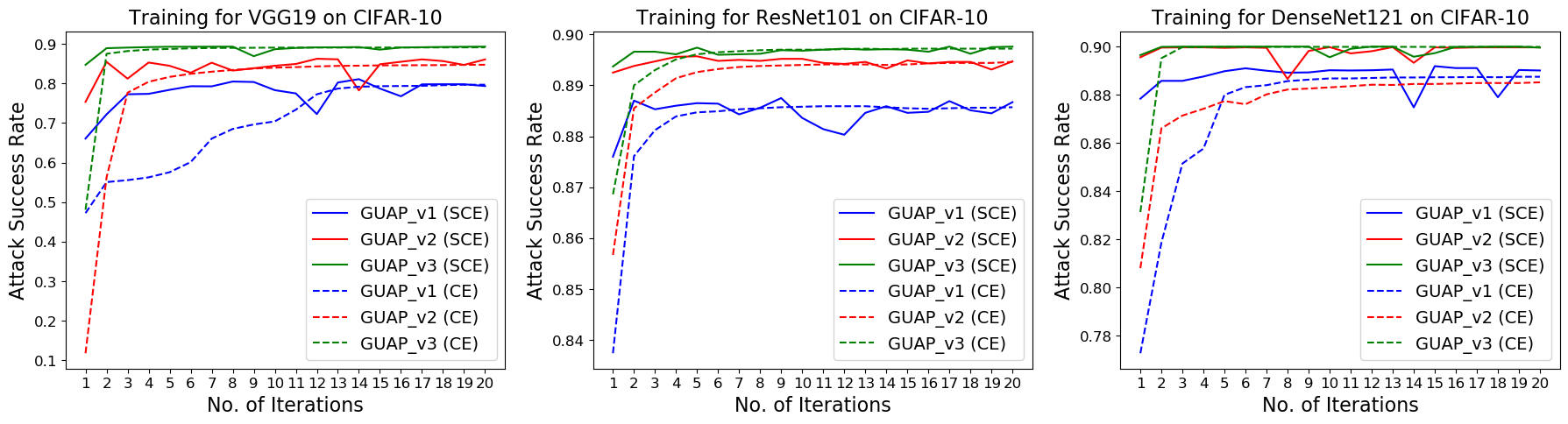}
\caption{Training process by maximizing the original cross-entropy loss and the scaled cross-entropy loss under different setups.} 
\label{sce}
\end{figure*}

\subsection{Effect of the Surrogate Loss Function}
In this part, we investigate the effect of scaled loss by comparing it with the performance of the original cross-entropy method on the CIFAR-10 dataset.
As shown in Fig.~\ref{sce}, it demonstrates the whole training process by applying the original cross-entropy loss (CE) and the scaled cross-entropy loss (SCE), respectively.
It can be observed that the performances achieved by the SCE loss function are better than obtained by CE loss function at the most time. Meanwhile, it also encourages faster convergence during training.

\section{Conclusion}
\label{conclusion}
In conclusion, we propose a unified framework for crafting universal adversarial perturbations, which can be either additive, non-additive, or a combination of both.
We show that, by combining non-additive perturbation such as spatial transformation with a small universal $\ell_\infty$-norm attack, our approach is able to obtain state-of-the-art attack success rate for universal adversarial attack, significantly outperforming existing approaches.
Moreover, comparing to current universal attacks, our approach can obtain higher fooling performance, yet with {\em i)} less amount of training data, {\em ii)} superior transferability of the attack, and {\em iii)} without compromising the human imperceptibility to adversarial examples. We believe, this work provides an alternative but more powerful universal adversarial attack solution, which marks a step forward to understand the global robustness of deep neural networks. 

\bibliographystyle{IEEEtran}
\bibliography{myicdm}

\end{document}